\RequirePackage[l2tabu, orthodox]{nag}
\RequirePackage{snapshot}

\documentclass[10pt,onecolumn]{article}

\makeatletter
\if@twocolumn
  \usepackage[dvips,letterpaper,top=0.5in, bottom=0.5in, left=0.75in, right=0.5in,includefoot,heightrounded]{geometry}
\else
  \usepackage[dvips,letterpaper,margin=1in,includefoot,heightrounded]{geometry}
\fi

\usepackage{srcltx}

\usepackage{amsmath}
\usepackage{amssymb,amsfonts}

\usepackage{abstract}

\usepackage{graphicx}
\usepackage{epsfig}
\usepackage[usenames,dvipsnames]{color}
\usepackage[usenames,dvipsnames]{xcolor}
\usepackage{subfigure}

\usepackage{booktabs}

\usepackage{setspace}
\usepackage{flushend}
\usepackage{multicol}

\usepackage{cite}
\usepackage{url}
\usepackage[normalem]{ulem}

\usepackage{enumerate}

\usepackage{array}

\usepackage{hyperref}

\usepackage{lipsum}

\usepackage[sc,tiny]{titlesec}

\usepackage{microtype}

\makeatletter

\newenvironment{rsmallmatrix}{\null\,\vcenter\bgroup
  \Let@\restore@math@cr\default@tag
  \baselineskip6\ex@ \lineskip1.5\ex@ \lineskiplimit\lineskip
  \ialign\bgroup\hfil$\m@th\scriptstyle##$&&\thickspace\hfil
  $\m@th\scriptstyle##$\crcr
}{%
  \crcr\egroup\egroup\,%
}
\makeatother

\title{%
Multiplierless 16-point DCT Approximation for Low-complexity Image and Video Coding
}

\author{%
T. L. T. Silveira%
\thanks{
T. L. T. Silveira
is with the
Programa de P\'os-Gradua\c{c}\~ao em Computa\c{c}\~ao, Universidade Federal do Rio Grande do Sul (UFRGS), Porto Alegre, RS, Brazil
}
\and
R. S. Oliveira%
\thanks{
R. S. Oliveira
is with the
Signal Processing Group,
Departamento de Estat\'{\i}stica,
Universidade Federal de Pernambuco (UFPE);
Programa de Gradua\c{c}\~ao em Estat\'{\i}stica (UFPE), Brazil,
and the Department of Electrical and Computer Engineering, University of Akron, OH
}
\and
F. M. Bayer%
\thanks{
F. M. Bayer
is with the
Departamento de Estat\'{\i}stica, UFSM, and LACESM, Santa Maria, RS, Brazil,
E-mail:~\protect\url{bayer@ufsm.br}
}
\and
R.~J.~Cintra%
\thanks{%
R.~J.~Cintra is with
the Signal Processing Group,
Departamento de Estat\'{\i}stica,
Universidade Federal de Pernambuco.
E-mail:~\protect\url{rjdsc@stat.ufpe.org}
}
\and
A. Madanayake%
\thanks{
A. Madanayake
is with the
Department of Electrical and Computer Engineering, University of Akron, OH,
E-mail:~\protect\url{arjuna@uakron.edu}
}
}

\date{}

\newcommand{\myabstract}{%
An orthogonal 16-point approximate discrete cosine transform (DCT) is introduced.
The proposed transform requires neither multiplications nor bit-shifting operations.
A fast algorithm based on matrix factorization is introduced, requiring only 44 additions---the
lowest arithmetic cost in literature.
To assess the introduced transform,
computational complexity,
similarity with the exact DCT,
and coding performance measures are computed.
Classical and state-of-the-art 16-point
low-complexity transforms
were used in a comparative analysis.
In the context of image compression,
the proposed approximation was evaluated via PSNR and SSIM
measurements,
attaining the best cost-benefit ratio among
the
competitors.
For video encoding,
the proposed approximation was embedded into a HEVC reference software
for direct comparison with the original HEVC standard.
Physically realized and tested using FPGA hardware,
the proposed transform
showed 35\% and 37\% improvements of area-time and
area-time-squared VLSI metrics
when compared to the
best competing transform in the literature.
}

\newcommand{\mykeywords}{%
DCT approximation,
Fast algorithm,
Low cost algorithms,
Image compression,
Video coding
}

\begin{document}

\makeatletter
\if@twocolumn

\twocolumn[%
  \maketitle
  \begin{onecolabstract}
    \myabstract
  \end{onecolabstract}
  \begin{center}
    \small
    \textbf{Keywords}
    \\\medskip
    \mykeywords
  \end{center}
  \bigskip
]
\saythanks

\else

  \maketitle
  \begin{abstract}
    \myabstract
  \end{abstract}
  \begin{center}
    \small
    \textbf{Keywords}
    \\\medskip
    \mykeywords
  \end{center}
  \bigskip
  \onehalfspacing
\fi

\section{Introduction}

The
discrete cosine transform (DCT)~\cite{britanak2007discrete,rao1990discrete}
is a fundamental building-block
for several image and video processing applications.
In fact,
the DCT closely approximates
the Karhunen-Lo\`eve transform (KLT)~\cite{britanak2007discrete},
which is capable of optimal
data decorrelation
and energy compaction
of first-order stationary Markov
signals~\cite{britanak2007discrete}.
This class of signals is particularly appropriate for
the modeling of
natural images~\cite{britanak2007discrete,cintra2014low}.
Thus,
the DCT finds
applications
in several
contemporary
image and video compression standards,
such as
the JPEG~\cite{penn1992}
and
the H.26x family of codecs~\cite{h261,h263,h2642003}.
Indeed,
several fast algorithms
for computing the exact DCT were
proposed~\cite{Chen1977,arai1988fast,fw1992,hou1987fast,lee1984new,loeffler1991practical,vetterli1984simple,wang1984fast}.
However,
these methods
require
the use of arithmetic
multipliers~\cite{lengwehasatit2004scalable,haweel2001new},
which are time, power, and hardware demanding
arithmetic operations,
when compared to additions or bit-shifting operations~\cite{blahut}.
This fact may jeopardize the application
of the DCT in very low power consumption contexts~\cite{tran,Lin2006}.
To overcome this problem,
in recent years,
several
approximate DCT methods
have been proposed.
Such approximations
do not compute the exact DCT,
but are capable of providing
energy compaction~\cite{bas2008,bayer201216pt}
at a very low computational cost.
In particular,
the 8-point DCT was given a number of approximations:
the signed DCT~\cite{haweel2001new},
the level~1~approximation~\cite{lengwehasatit2004scalable},
the Bouguezel-Ahmad-Swamy~(BAS) transforms~\cite{bas2008,bas2009,bas2010,bas2011,bas2013},
the rounded DCT (RDCT)~\cite{cb2011},
the modified RDCT~\cite{bc2012},
the approximation in~\cite{multibeam2012},
and
the improved DCT approximation introduced in~\cite{Potluri2013}.
These methods
furnish meaningful DCT approximations
using only
addition and bit-shifting operations,
whilst offering sufficient computational
accuracy for image and video processing~\cite{mssp2014}.

Recently,
with the growing need for higher compression rates~\cite{Potluri2013},
the high efficiency video coding~(HEVC)
was proposed~\cite{hevc1,Sullivan2012}.
Unlike several image and video compression standards,
the HEVC employs
4-, 16-, and 32-point integer DCT-based transformations~\cite{hevc1,Potluri2013}.
In contrast to
the 8-point DCT case---where
dozens of approximations
are
available~\cite{bas2008,bouguezel2008multiplication,bas2011,bc2012,Potluri2013,cb2011},
---the 16-point DCT approximation methods
are
much less explored in literature.
To the best of our knowledge,
only the following orthogonal methods are available:
the traditional Walsh--Hadamard transform (WHT)~\cite{yarlagadda},
the BAS-2010~\cite{bas2010} and BAS-2013~\cite{bas2013}
approximations,
and
the transformations
proposed in~\cite{mssp2014},~\cite{bayer201216pt}, and~\cite{Jridi2015}.

In this work,
we aim at proposing
a low-complexity orthogonal 16-point DCT approximation
capable of outperforming
all competing methods in terms of
arithmetic complexity
while
exhibiting very close coding performance
when compared to state-of-the-art methods.
For such,
we advance
a transformation matrix which combines
instantiations of a low-complexity 8-point approximation
according to a divide-and-conquer approach.

The remainder of this paper is organized as follows.
Section~\ref{sec:methodology} introduces the new DCT approximation,
a fast algorithm based on matrix factorization,
and
a comprehensive  assessment
in terms of
computational complexity
and several performance metrics.
In Section~\ref{sec:imageandvideocompression},
the proposed approximation
is submitted to computational simulations
consisting
of
a JPEG-like scheme for still image compression
and
the embedding of the proposed approximation
into a HEVC standard reference software.
Section~\ref{section-hardware}
assesses the proposed transform
in a hardware realization based on
field-programmable gate array (FPGA).
Conclusions are drawn in Section~\ref{sec:conclusion}.

\section{16-point DCT approximation}
\label{sec:methodology}

\subsection{Definition}
\label{sec:definition}

\begin{figure}%
\centering
\scalebox{1}{\input{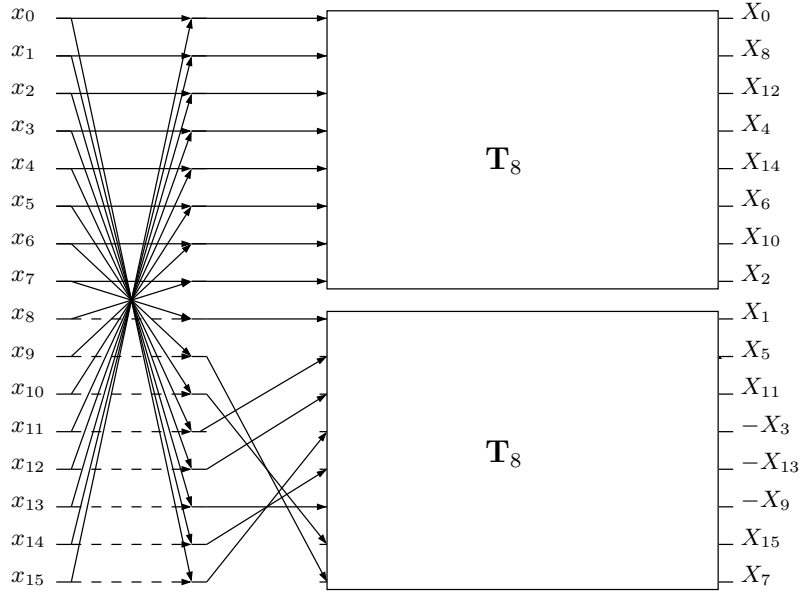}}
\caption{Signal-flow graph of the fast algorithm for $\mathbf{T}$. The input data $x_i$,
$i = 0,1,\ldots,15$ relates to the output data $X_j$,
$j = 0,1,\ldots,15$ according to $\mathbf{X}=\mathbf{T}\cdot\mathbf{x}$. Dashed arrows represent multiplications by -1.}
\label{f:grafo}
\end{figure}

It is well-known that several fast algorithm structures
compute the $N$-point DCT
through recursive
computations of the $\frac{N}{2}$-point DCT~\cite{mssp2014, britanak2007discrete, loeffler1991practical, rao1990discrete, Jridi2015}.
Following a similar approach to that adopted in~\cite{mssp2014,Jridi2015},
we propose a new 16-point approximate DCT
by
combining
two instantiations of the 8-point DCT approximation
introduced in~\cite{bc2012}
with
tailored
signal changes and permutations.
This procedure is induced by signal-flow graph in Fig.~\ref{f:grafo}.
This particular 8-point DCT approximation,
presented as $\mathbf{T}_8$ in Fig.~\ref{f:grafo},
was selected because
(i)~it presents the lowest computational cost
among the approximations archived in literature
(zero multiplications,
14~additions,
and
zero bit-shifting operations)~\cite{bc2012}
and
(ii)~it offers good energy compaction properties~\cite{Tablada2015}.

As a result,
the proposed transformation matrix
is given by:
\begin{align*}
\mathbf{T}
=
\left[
\begin{rsmallmatrix}
1 & 1 & 1 & 1 & 1 & 1 & 1 & 1 & 1 & 1 & 1 & 1 & 1 & 1 & 1 & 1\\
1 & 1 & 1 & 1 & 1 & 1 & 1 & 1 & -1 & -1 & -1 & -1 & -1 & -1 & -1 & -1\\
1 & 0 & 0 & 0 & 0 & 0 & 0 & -1 & -1 & 0 & 0 & 0 & 0 & 0 & 0 & 1\\
1 & 1 & 0 & 0 & 0 & 0 & -1 & -1 & 1 & 1 & 0 & 0 & 0 & 0 & -1 & -1\\
1 & 0 & 0 & -1 & -1 & 0 & 0 & 1 & 1 & 0 & 0 & -1 & -1 & 0 & 0 & 1\\
1 & 1 & -1 & -1 & -1 & -1 & 1 & 1 & -1 & -1 & 1 & 1 & 1 & 1 & -1 & -1\\
0 & 0 & -1 & 0 & 0 & 1 & 0 & 0 & 0 & 0 & 1 & 0 & 0 & -1 & 0 & 0\\
0 & 0 & 0 & 0 & 0 & 0 & -1 & 1 & -1 & 1 & 0 & 0 & 0 & 0 & 0 & 0\\
1 & -1 & -1 & 1 & 1 & -1 & -1 & 1 & 1 & -1 & -1 & 1 & 1 & -1 & -1 & 1\\
0 & 0 & -1 & 1 & 0 & 0 & 0 & 0 & 0 & 0 & 0 & 0 & -1 & 1 & 0 & 0\\
0 & -1 & 0 & 0 & 0 & 0 & 1 & 0 & 0 & 1 & 0 & 0 & 0 & 0 & -1 & 0\\
0 & 0 & 1 & 1 & -1 & -1 & 0 & 0 & 0 & 0 & 1 & 1 & -1 & -1 & 0 & 0\\
0 & -1 & 1 & 0 & 0 & 1 & -1 & 0 & 0 & -1 & 1 & 0 & 0 & 1 & -1 & 0\\
1 & -1 & 0 & 0 & 0 & 0 & 0 & 0 & 0 & 0 & 0 & 0 & 0 & 0 & 1 & -1\\
0 & 0 & 0 & -1 & 1 & 0 & 0 & 0 & 0 & 0 & 0 & 1 & -1 & 0 & 0 & 0\\
0 & 0 & 0 & 0 & -1 & 1 & 0 & 0 & 0 & 0 & -1 & 1 & 0 & 0 & 0 & 0
\end{rsmallmatrix}
\right]
.
\end{align*}
The entries
of the resulting
transformation matrix
are
defined over
$\{0,\pm1\}$,
therefore
it is completely multiplierless.
Above transformation can
be
orthogonalized
according to the procedure described
in~\cite{cintra2011integer,cb2011,cintra2014low}.
Thus
the associate orthogonal DCT approximation
is furnished
by
$\mathbf{\hat{C}} = \mathbf{S} \cdot \mathbf{T}$,
where
$
\mathbf{S} = \sqrt{(\mathbf{T}\cdot\mathbf{T}^\top)^{-1}}
$
and
the superscript~${}^\top$
denotes matrix transposition.
In particular,
we have:
\begin{align*}
 \mathbf{S}
=
&
\frac{1}{4}
\cdot
\operatorname{diag}
\left(
1,
1,
2,
\sqrt{2},
\sqrt{2},
1,
2,
2,
1,
2,
2,
\sqrt{2},
\sqrt{2},
2,
2,
2
\right)
.
\end{align*}
In the context of image and video coding,
the diagonal matrix~$\mathbf{S}$
does not contribute to
the computational cost of~$\mathbf{\hat{C}}$.
This is because it can be merged into
the codec quantization steps~\cite{mssp2014,bayer201216pt,bas2011,cb2011}.
Therefore,
the actual computation cost of the
approximation
is
fully confined in the
low-complexity matrix~$\mathbf{T}$.

\subsection{Fast algorithm and computational complexity}
\label{sec:evaluation}

The transformation $\mathbf{T}$ requires 112~additions,
if computed directly.
However,
it can be given the following
sparse matrix factorization:
\begin{align*}
\mathbf{T}
=
\mathbf{P}_2
\cdot
\mathbf{M}_4
\cdot
\mathbf{M}_3
\cdot
\mathbf{M}_2
\cdot
\mathbf{P}_1
\cdot
\mathbf{M}_1
,
\end{align*}
where
\begin{align*}
\mathbf{M}_1 &=
\left[
\begin{rsmallmatrix}\\
\mathbf{I}_8 & \overline{\mathbf{I}}_8\\
\overline{\mathbf{I}}_8 & -\mathbf{I}_8\\
\end{rsmallmatrix}
\right]
,
\\
\mathbf{M}_2 &= \operatorname{diag}\left(
 \left[
 \begin{rsmallmatrix}\\
 \mathbf{I}_4 & \overline{\mathbf{I}}_4\\
 \overline{\mathbf{I}}_4 & -\mathbf{I}_4\\
 \end{rsmallmatrix} \right],
 \left[
 \begin{rsmallmatrix}\\
 \mathbf{I}_4 & \overline{\mathbf{I}}_4\\
 \overline{\mathbf{I}}_4 & -\mathbf{I}_4\\
 \end{rsmallmatrix} \right]
 \right)
,
\\
\mathbf{M}_3 &= \operatorname{diag}\left(
  \left[
 \begin{rsmallmatrix}\\
 \mathbf{I}_2 & \overline{\mathbf{I}}_2\\
 \overline{\mathbf{I}}_2 & -\mathbf{I}_2\\
 \end{rsmallmatrix} \right],
  -\mathbf{I}_4,
  \left[
 \begin{rsmallmatrix}\\
 \mathbf{I}_2 & \overline{\mathbf{I}}_2\\
 \overline{\mathbf{I}}_2 & -\mathbf{I}_2\\
 \end{rsmallmatrix} \right]
 , -\mathbf{I}_4
 \right)
,
\\
\mathbf{M}_4 &= \operatorname{diag}\left(
 \left[
 \begin{rsmallmatrix}\\
 1 & 1 & 0\\
 1 & -1 & 0\\
 0 & 0 & -1\\
 \end{rsmallmatrix}
 \right],
\mathbf{I}_4,
 \left[
 \begin{rsmallmatrix}\\
 -1 & 0 & 0\\
 0 & 1 & 1\\
 0 & 1 & -1\\
 \end{rsmallmatrix}
 \right],
 -\mathbf{I}_4,
 \left[
 \begin{rsmallmatrix}\\
 1 & 0\\
 0 & -1\\
 \end{rsmallmatrix}
 \right]
 \right)
,
\end{align*}
matrices
$\mathbf{P}_1$ and $\mathbf{P}_2$ correspond
to the permutations
(1)(2)(3)(4)(5)(6)(7)(8)(9)(10 12 16 10)(11 13 15 11)(14)
and
(1)(2 9)(3 8 16 15 5 4 12 11 7 6 10 14 13 3)
in cyclic notation~\cite{Herstein1975},
respectively;
and
$\mathbf{I}_N$ and $\overline{\mathbf{I}}_N$
denote the identity and counter-identity matrices
of order $N$,
respectively.
The above factorization
reduces the computational cost
of~$\mathbf{T}$
to only 44~additions.
Fig.~\ref{f:grafo}
depicts the signal-flow graph
of the fast algorithm for $\mathbf{T}$;
the blocks labeled as
$\mathbf{T}_8$
denote the selected 8-point approximate DCT~\cite{bc2012}.

A computational complexity comparison
of the considered
orthogonal 16-point DCT approximations is summarized in
Table~\ref{tab:complexity}.
For contrast,
we also included
the computational cost of
the Chen DCT fast algorithm~\cite{Chen1977}.
The proposed approximation requires
neither multiplication, nor bit-shifting operations.
Furthermore,
when compared to
the methods in~\cite{mssp2014,Jridi2015},
the WHT or BAS-2013,
and the transformation in~\cite{bayer201216pt},
the proposed approximation
requires
26.67\%,
31.25\%,
and
38.89\%
less arithmetic operations,
respectively.

\begin{table}%
\centering
\caption{Comparison of computational complexities}
\label{tab:complexity} %
\begin{tabular}{l|c|c|c|c} %
\toprule
Transform		 & Mult & Add &  Shifts  & Total  \\
\midrule
Chen DCT 	 		& 44 & 74 	& 0	  	& 118 \\
WHT 	 		& 0 & 64 	& 0	  	& 64 \\
BAS-2010	 	& 0 & 64 	&  8 			&72\\
BAS-2013 		& 0 & 64  		 & 0 	&64\\
Transform in~\cite{bayer201216pt}		& 0 & 72 	& 0	  	& 72 \\
Transform in~\cite{mssp2014}		& {0} & {60} 	& {0}	  	& {60}\\
Transform in~\cite{Jridi2015} &0 & 60 & 0 & 60\\
\textbf{Proposed approx.}          & \textbf{0}   & \textbf{44}       &  \textbf{0}	            &\textbf{44}\\
\bottomrule
\end{tabular}
\end{table}

\subsection{Performance assessment}

We separate similarity and coding performance measures
to assess the
proposed
transformation.
For similarity measures,
we considered
the DCT distortion ($d_2$)~\cite{fong2012},
the total error energy~($\epsilon$)~\cite{cb2011},
and
the mean square error (MSE)~\cite{britanak2007discrete,rao1990discrete}.
For coding performance evaluation,
we selected the
the transform coding gain~($C_g$)~\cite{britanak2007discrete}
and
the
transform efficiency~($\eta$)~\cite{britanak2007discrete}.
Table~\ref{tab:performances}
compares
the performance measure values for
the discussed transforms.
The proposed approximation
could furnish performance measure
which are comparable to the average results
of the state-of-the-art approximation.
At the same time,
its computational cost is roughly 30\%~smaller
than the lowest complexity method in literature~\cite{mssp2014,Jridi2015}.

\begin{table}%

\centering

\caption{Coding and similarity performance assessment}
\label{tab:performances}
\begin{tabular}{l|p{.6cm}|p{.8cm}|p{.7cm}|p{.6cm}|p{.6cm}}
\toprule
Transform& $d_2$ & $\epsilon$ & MSE & $C_g$ & $\eta$ \\
\midrule
Chen DCT & 0.000 & 0.000 & 0.000 &  9.455& 88.452  \\
WHT &0.878& 92.563 & 0.428 &  8.194 & 70.646  \\
BAS-2010&0.667	 & 64.749 &  0.187 & 8.521 & 73.634 	\\
BAS-2013&0.511	 & 54.621  & 0.132 & 8.194 & 70.646	\\
Transform in~\cite{bayer201216pt}&0.152	& 8.081 & 0.046 & 7.840 & 65.279 \\
Transform in~\cite{mssp2014}&0.340& 30.323& 0.064 &  8.295& 70.831 \\
Transform in~\cite{Jridi2015} &0.256 & 14.740 & 0.051 & 8.428 & 72.230 \\
\textbf{Proposed approx.}  & \textbf{0.493} & \textbf{41.000}  &  \textbf{0.095}&\textbf{7.857} &\textbf{67.608}\\
\bottomrule

\end{tabular}
\end{table}

\section{Image and video coding}
\label{sec:imageandvideocompression}

In the following subsections,
we describe two computational experiments
in the context of image and video encoding.
Our goal is to demonstrate in real-life scenarios
that the introduced approximation
is capable of performing
very closely to state-of-the-art
approximations
at a much lower computational cost.
For the still image
experiment,
we employ a fixed-rate encoding scheme
which avoids quantization.
This is done to isolate the role of the transform
in order to emphasize the good properties of energy compaction
of the approximate transforms.
On the other hand,
for the video experiment,
we include
the variable-rate encoding
equipped with the quantization step
as required by the actual HEVC standard.
Thus,
we aim at providing two comprehensive experiments
to highlight the capabilities of the introduced
approximation.

\subsection{Image compression experiments}
\label{sec:imagecompression}

\begin{figure}%
\centering
\subfigure[PSNR]
{\includegraphics[width=0.46\textwidth]{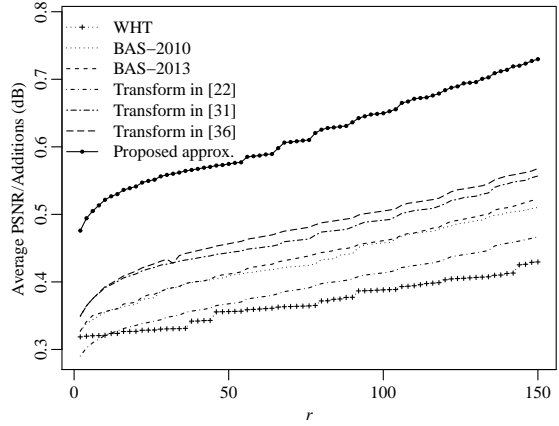}
\label{f:psnr}}
\\
\subfigure[SSIM]
{\includegraphics[width=0.46\textwidth]{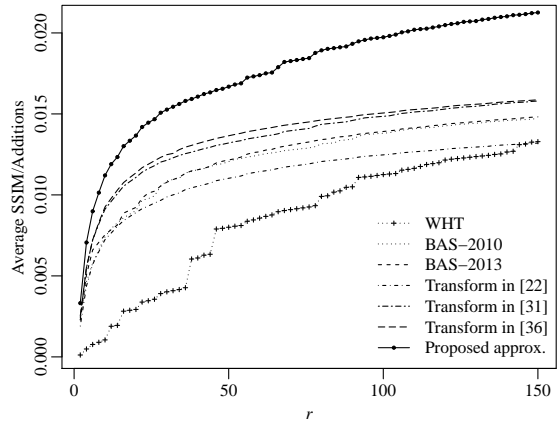}
\label{f:ssim}}
\caption{
Average
\subref{f:psnr} PSNR and
\subref{f:ssim} SSIM measurements
per additive cost
at
compression ratios.}
\label{fig:averagemeasures}
\end{figure}

We adopted a JPEG-like procedure
as
detailed in
the methodology presented in~\cite{haweel2001new}
and
reproduced in~
\cite{bas2008, bas2010, bas2011, Jridi2015, mssp2014}.
A total of 45 512$\times$512 8-bit grayscale images obtained from a standard public image bank~\cite{uscsipi} was considered.
This set of image was selected to be representative
of the imagery commonly found in real-life applications.
Color images could be treated similarly by processing each channel separately.
Each given input image $\mathbf{A}$
was split into 1024~16$\times$16
disjoint blocks
($\mathbf{A}_k$, $k=1,2,\ldots, 1024$)
which
were submitted to the forward bidimensional (\mbox{2-D})
transformation given by:
$
\mathbf{B}_k
=
\mathbf{\tilde{C}}
\cdot
\mathbf{A}_k
\cdot
\mathbf{\tilde{C}^\top}
$,
where $\mathbf{\tilde{C}}$ is a selected
16-point transformation.
Following the zig-zag sequence~\cite{pao1998},
only the first $1\leq r \leq 150$
elements
of
$\mathbf{B}_k$
were retained;
being the remaining ones zeroed
and resulting
in~$\tilde{\mathbf{B}}_k$.
The inverse \mbox{2-D} transformation is then applied
according to:
$
\mathbf{\tilde{A}}_k
=
\mathbf{\tilde{C}^\top}
\cdot
\tilde{\mathbf{B}}_k
\cdot
\mathbf{\tilde{C}}
$.
The resulting matrix $\mathbf{\tilde{A}}_k$
is the lossy
reconstruction of $\mathbf{A}_k$.
The correct rearrangement of all blocks
results in the reconstructed image $\mathbf{\tilde A}$.
This procedure was performed for each of the 45~images
in the selected data set.
To assess the approximation
in a fair manner,
we consider
the ratio
between
performance measures
and
arithmetic cost.
Such ratio
furnishes
the performance gain
per unit
of arithmetic computation.
Fig.~\ref{fig:averagemeasures} shows
the
average PSNR
and
structural similarity index~(SSIM)~\cite{Wang2004} measurements
per unit of additive cost.
The
proposed approximation
outperforms
all approximate DCT
for
any value of $r$ in both metrics.
The introduced 16-point transform
presents the best cost-benefit ratio
among all competing methods.

\begin{figure}%
\centering

\subfigure[Original image]
{\includegraphics[width=0.25\textwidth]{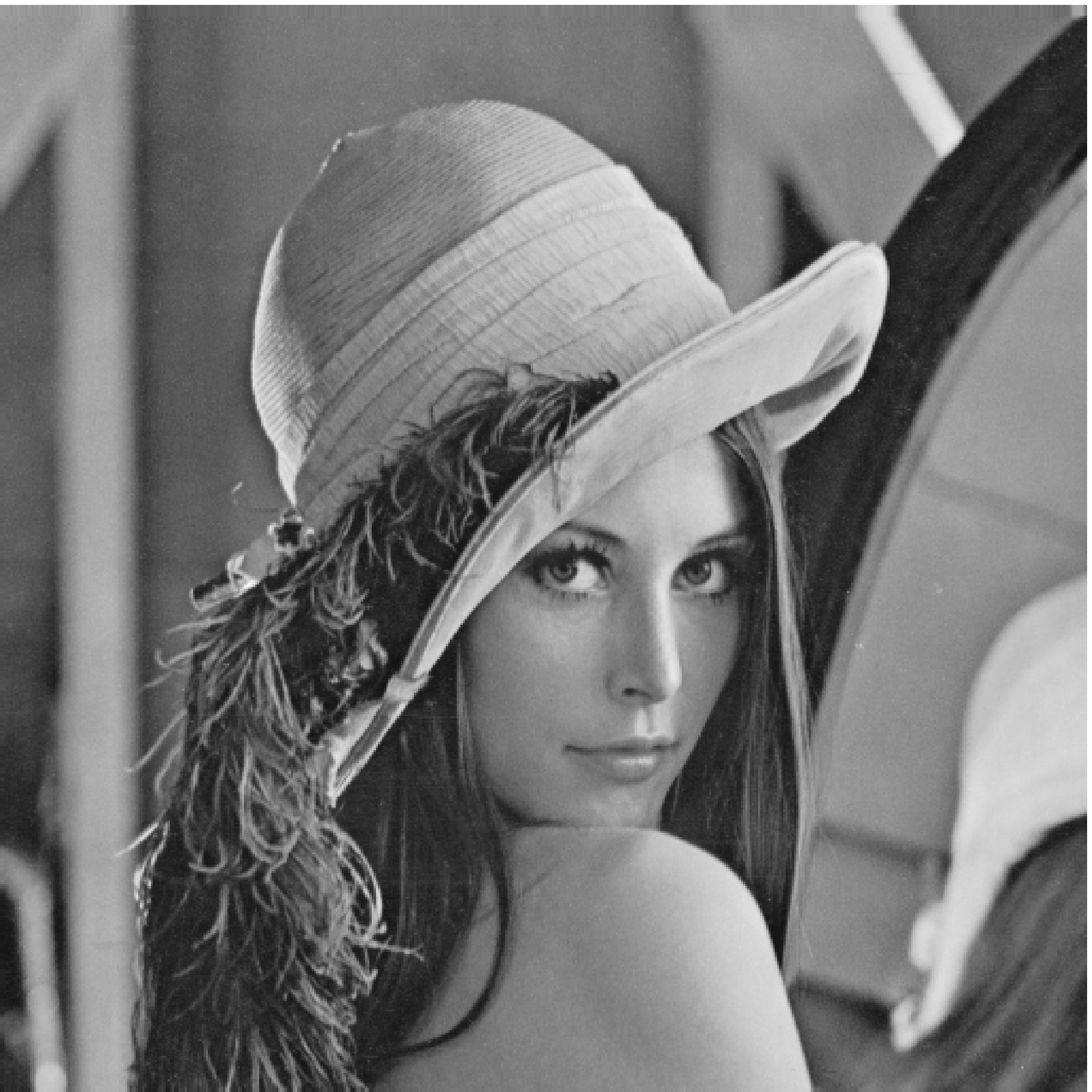}
\label{f:lena}
}\
\subfigure[$\text{PSNR}=28.55~\mathrm{dB}$, $\text{SSIM}=0.7915$]
{\includegraphics[width=0.25\textwidth]{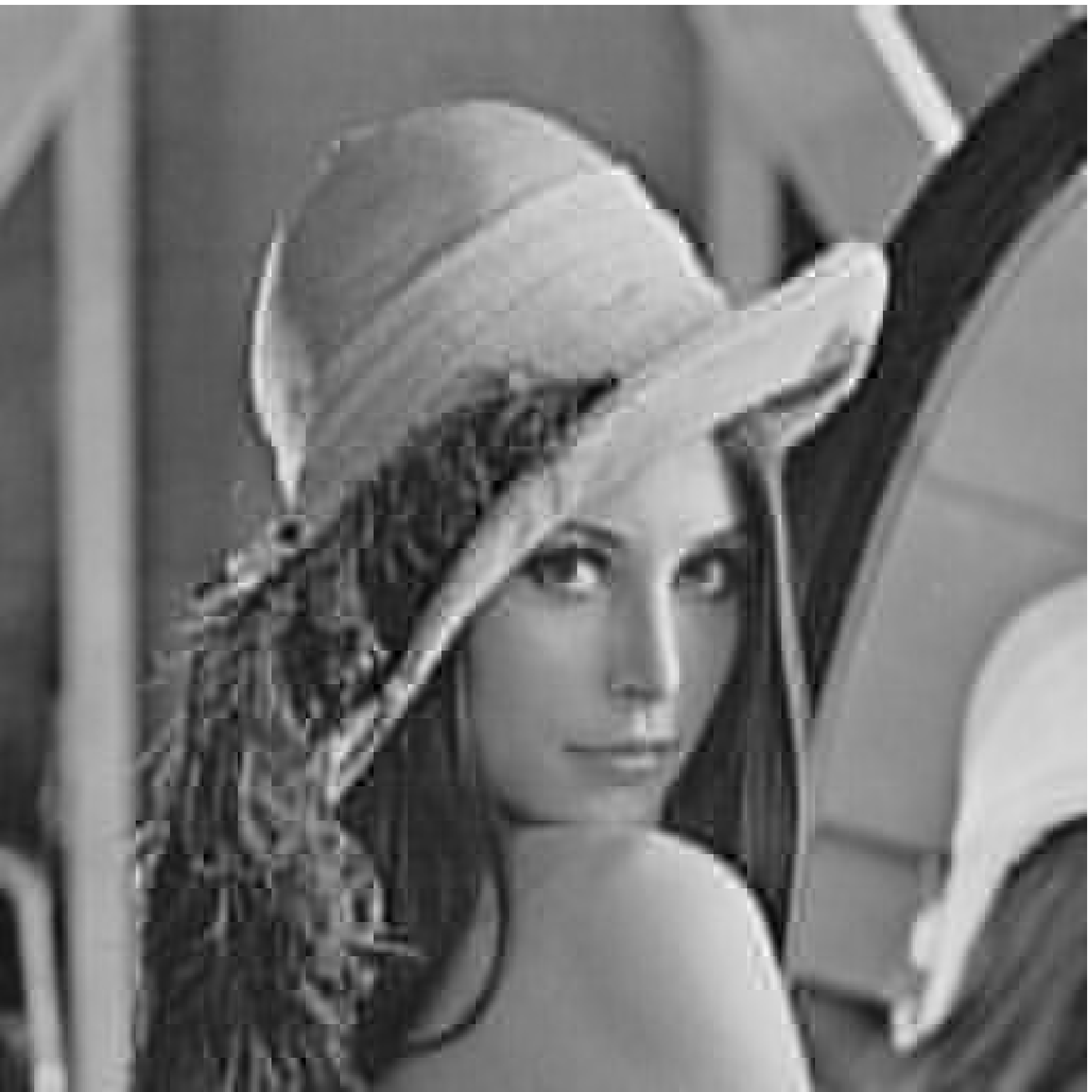}
\label{f:lenadct}
}\
\subfigure[$\text{PSNR}=21.20~\mathrm{dB}$, $\text{SSIM}=0.2076$]
{\includegraphics[width=0.25\textwidth]{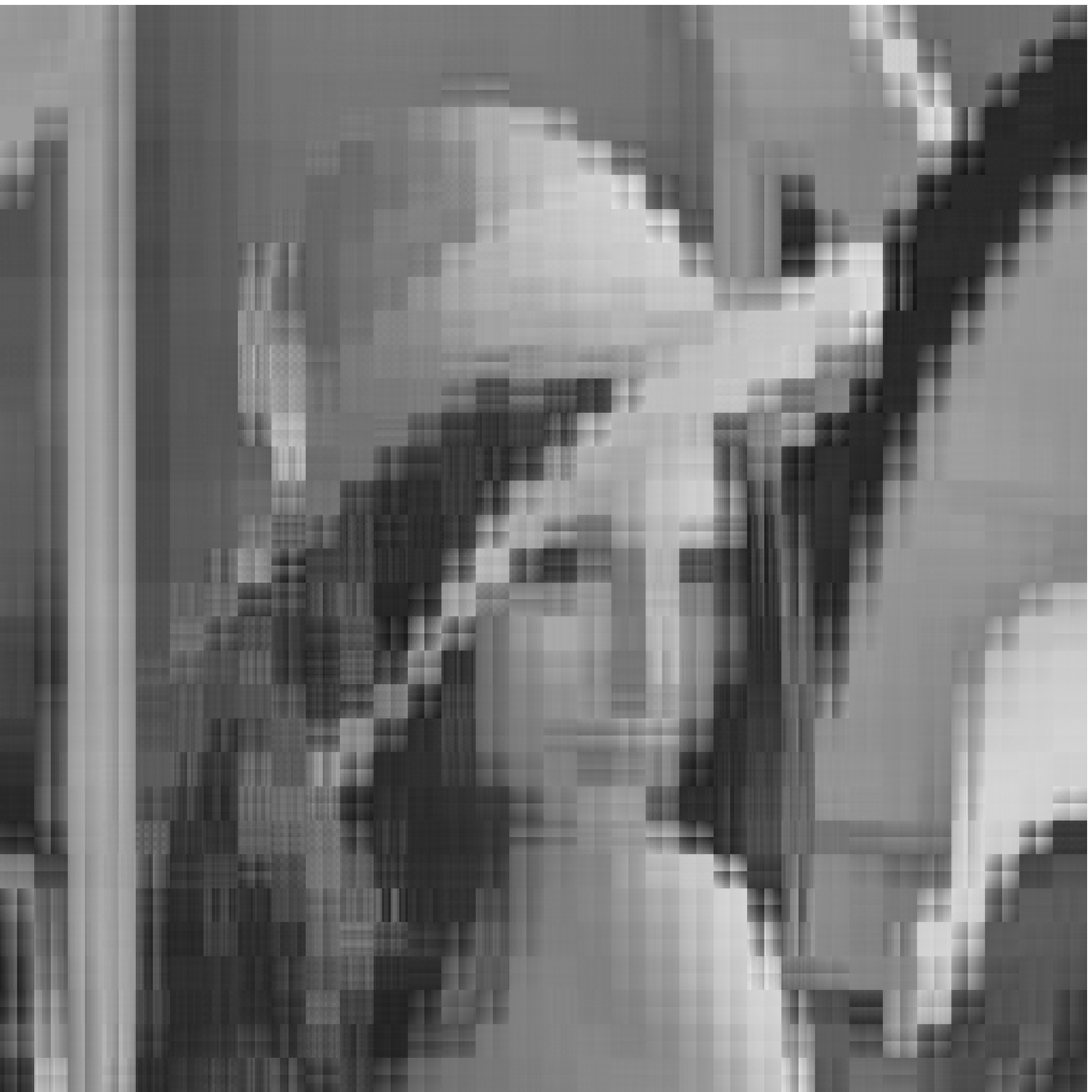}
\label{f:lenawht}
}\\
\subfigure[$\text{PSNR}=25.27~\mathrm{dB}$, $\text{SSIM}=0.6735$]
{\includegraphics[width=0.25\textwidth]{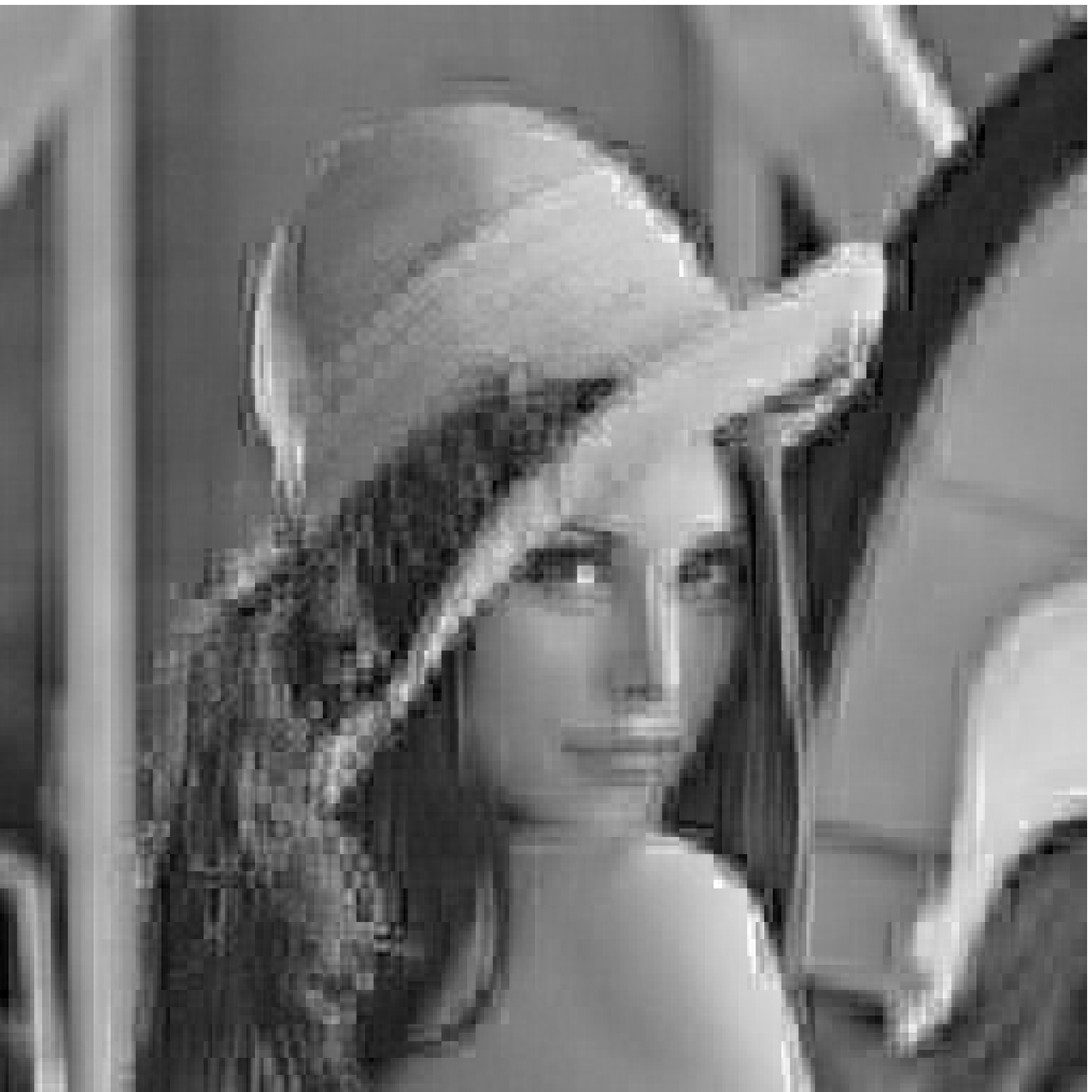}
\label{f:lenabas10}}\
\subfigure[$\text{PSNR}=25.79~\mathrm{dB}$, $\text{SSIM}=0.6921$]
{\includegraphics[width=0.25\textwidth]{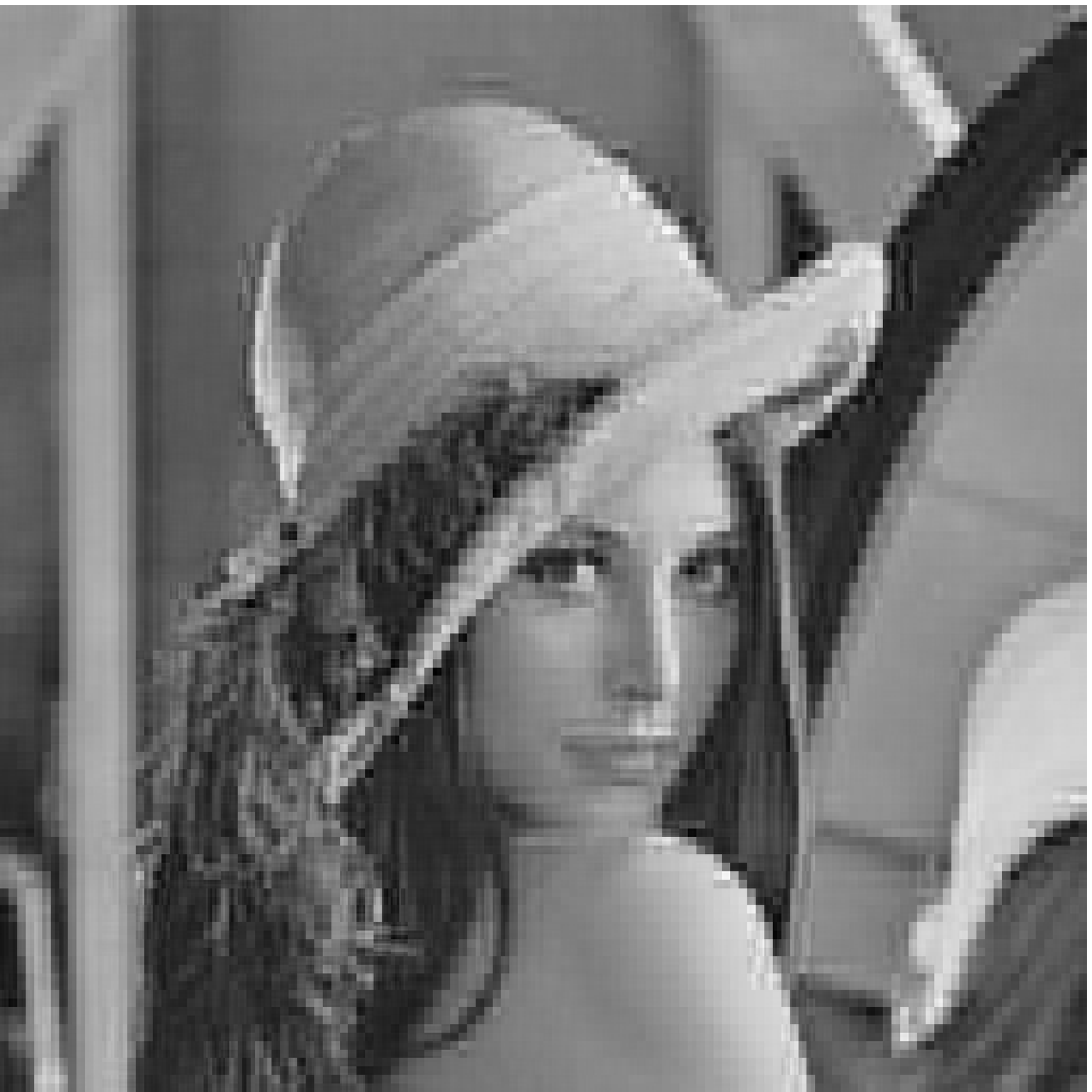}
\label{f:lenabas13}}\
\subfigure[$\text{PSNR}=25.75~\mathrm{dB}$, $\text{SSIM}=0.7067$]
{\includegraphics[width=0.25\textwidth]{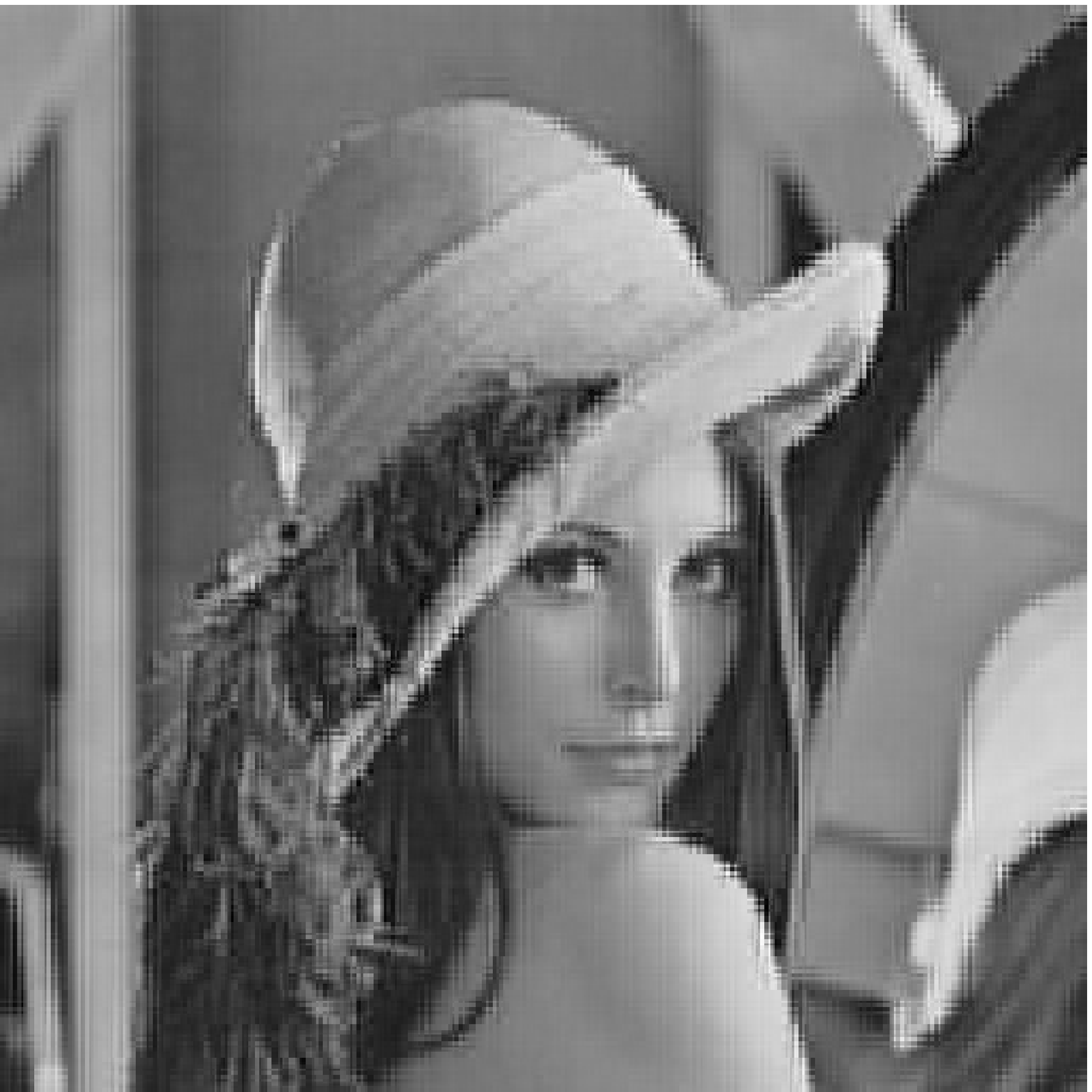}
\label{f:lenabcem}}\\
\subfigure[$\text{PSNR}=27.13~\mathrm{dB}$, $\text{SSIM}=0.7505$]
{\includegraphics[width=0.25\textwidth]{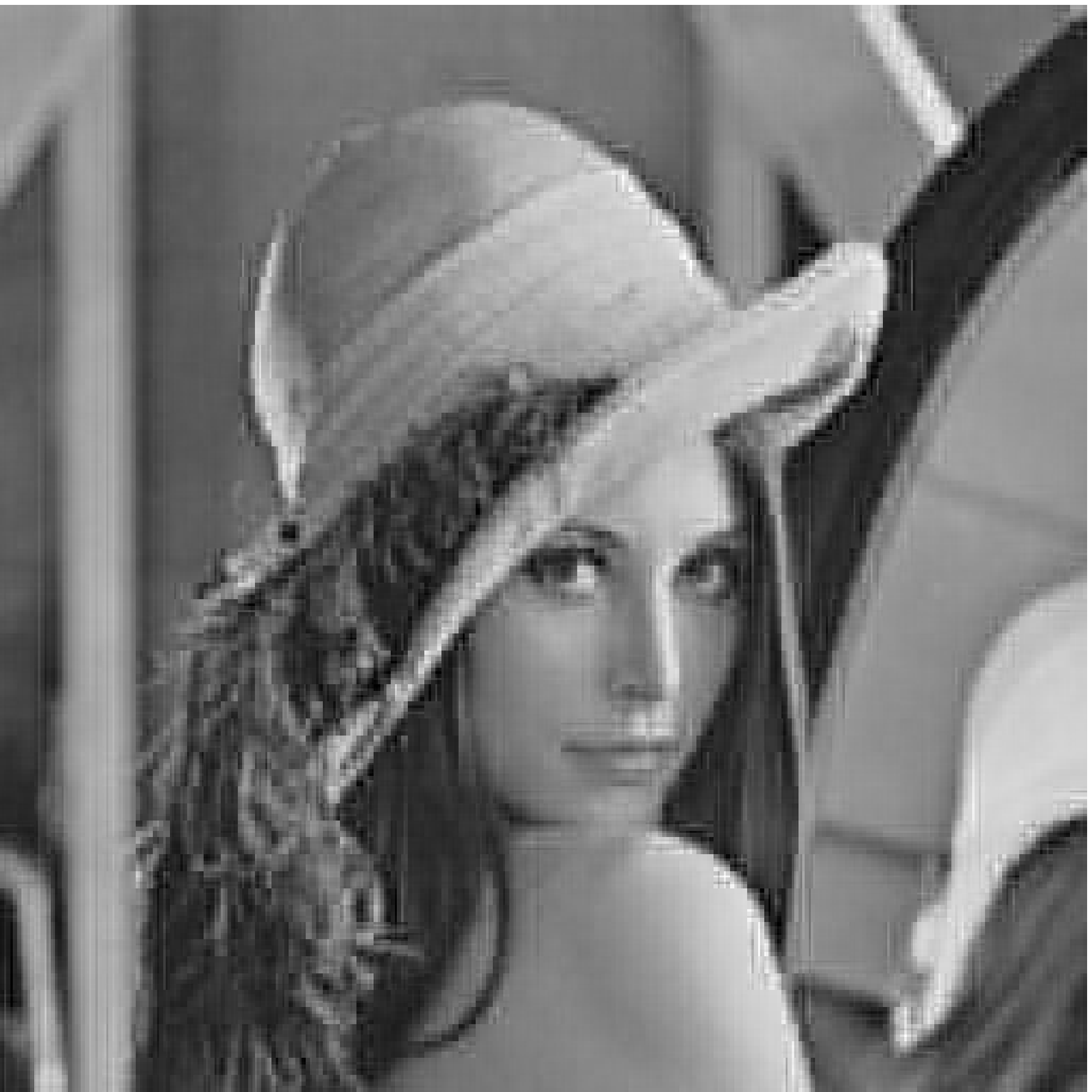}
\label{f:lenasbckmk}}\
\subfigure[$\text{PSNR}=27.40~\mathrm{dB}$, $\text{SSIM}=0.7587$]
{\includegraphics[width=0.25\textwidth]{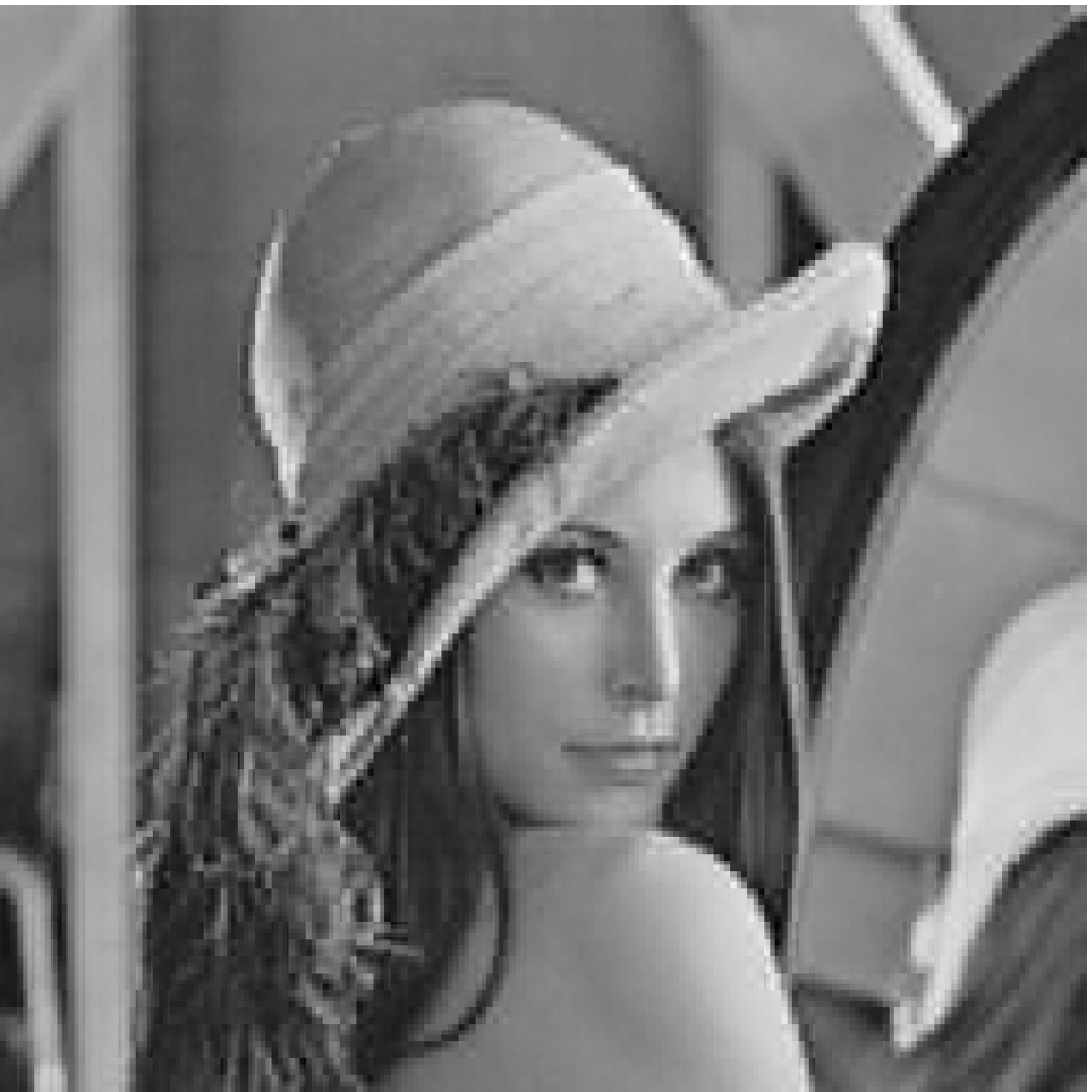}
\label{f:lenajridi}}\
\subfigure[$\text{PSNR}=25.84~\mathrm{dB}$, $\text{SSIM}=0.7023$]
{\includegraphics[width=0.25\textwidth]{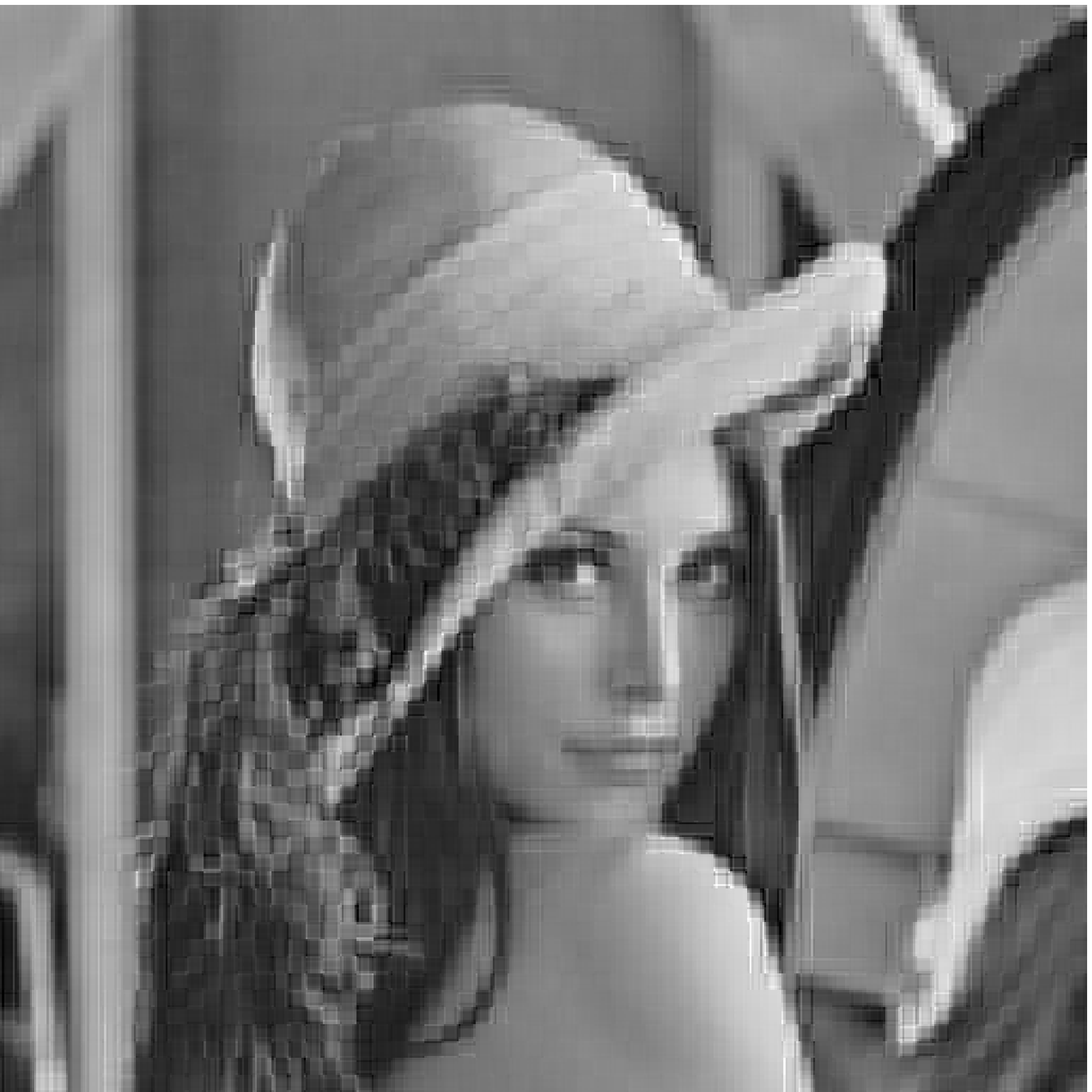}
\label{f:lenanew}}

\caption{%
Original
(a)~Lena image and compressed versions with $r=16$
according to
(b)~the DCT,
(c)~WHT,
(d)~BAS-2010,
(e)~BAS-2013,
(f)~transform in \cite{bayer201216pt},
(g)~transform in \cite{mssp2014},
(h)~transform in \cite{Jridi2015},
and
(i)~proposed 16-point approximation.}
\label{f:alllena}
\end{figure}

Fig.~\ref{f:alllena}
displays
a qualitative and quantitative comparison
considering standard Lena image.
The PSNR measurements for
the Lena image
were
only 4.75\% and 5.69\% below
the results furnished by the transformations
in~\cite{mssp2014,Jridi2015},
respectively.
Similarly,
considering the SSIM,
the proposed transform performed
only 0.62\%, 6.42\%, and 7.43\% below
the performance offered by the transformations
in~\cite{bayer201216pt}, \cite{mssp2014}, and~\cite{Jridi2015}.
On the other hand,
the proposed approximate DCT
requires
38.8\%
and
26.6\%
less arithmetic operations
when compared to~\cite{bayer201216pt}
and~\cite{mssp2014,Jridi2015},
respectively.
The proposed approximation
outperformed
the WHT, BAS-2010, and BAS-2013 according to both figures of merit.
Indeed,
the small losses in PSNR and SSIM
compared to
the exact DCT
are not sufficient
to effect
a significant image degradation
as perceived by
the human visual system,
as shown in
Fig.~\ref{f:alllena}.

\subsection{Video compression experiments}
\label{sec:videocompression}

The proposed approximation
was
embedded into the HM-16.3 HEVC reference software~\cite{refsoft},
i.e.,
the proposed approximation
is
considered as a replacement
for the original integer
transform
in the HEVC standard.
Because
the HEVC standard employs 4-, \mbox{8-,} \mbox{16-,} and 32-point
transformations,
we performed simulations in two scenarios:
(i)~substitution of the 16-point transformation only
and
(ii)~replacement of the 8- and 16-point transformations.
We adopted
the approximation described in~\cite{bc2012}
and the proposed approximation
for the 8- and 16-point substitutions,
respectively.
The original 8- and 16-point transforms
employed in the HEVC standard
require 22~multiplications and 28~additions;
and
86~multiplications and 100~additions,
respectively~\cite{Budagavi2012}.
In contrast,
the selected DCT approximations
are
multiplierless
and
require
50\%
and
56\%
fewer additions,
respectively.
The diagonal matrices associated to the 8- and 16-point approximations are
fully embedded into the quantization step
according to judicious scaling operations of the standard HEVC
quantization tables~\cite{Budagavi2012}.

\begin{figure}%
\centering
{\includegraphics[width=0.42\textwidth]{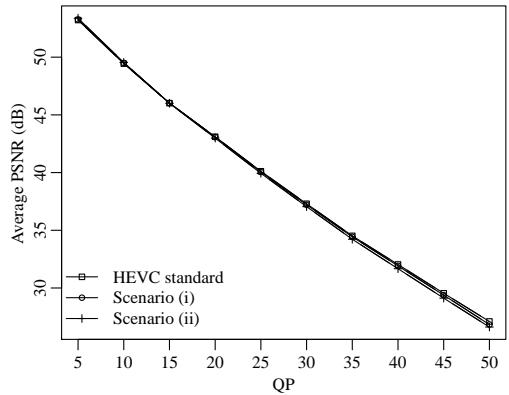}
\label{f:psnrhevc}}
\vspace{-0.3cm}
\caption{Performance of the proposed DCT approximation in HEVC standard for several QP values.}
\label{fig:graphicshevc}
\end{figure}

\begin{figure}%
\centering
\subfigure[HEVC standard]
{\includegraphics[width=0.25\textwidth]{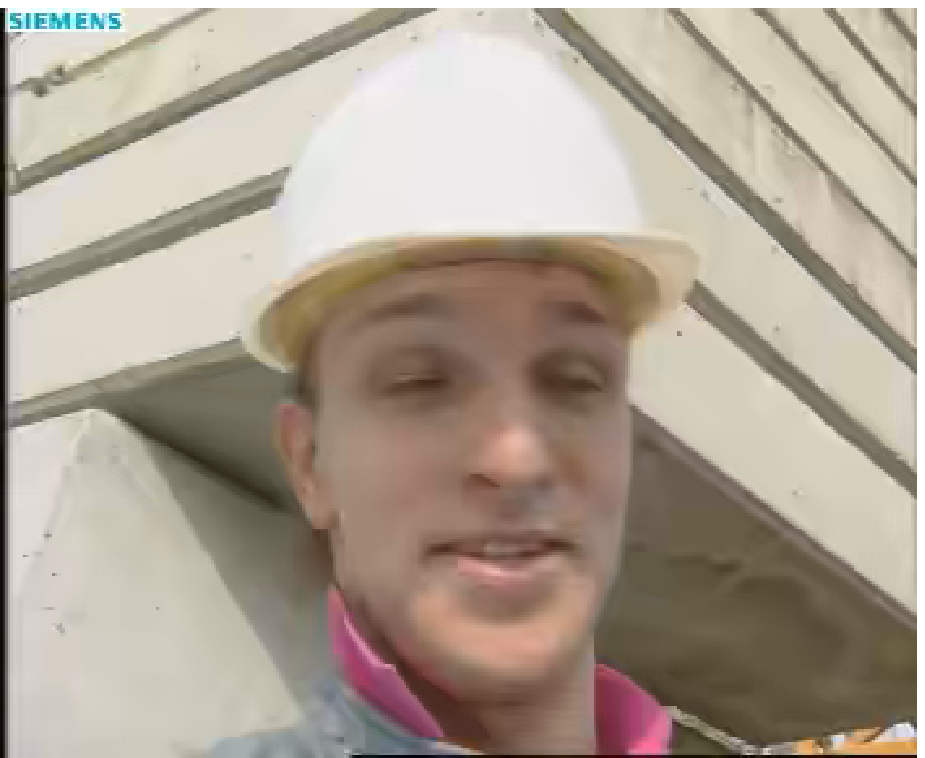}
\label{f:foremanhevc}}
\subfigure[Scenario (i)]
{\includegraphics[width=0.25\textwidth]{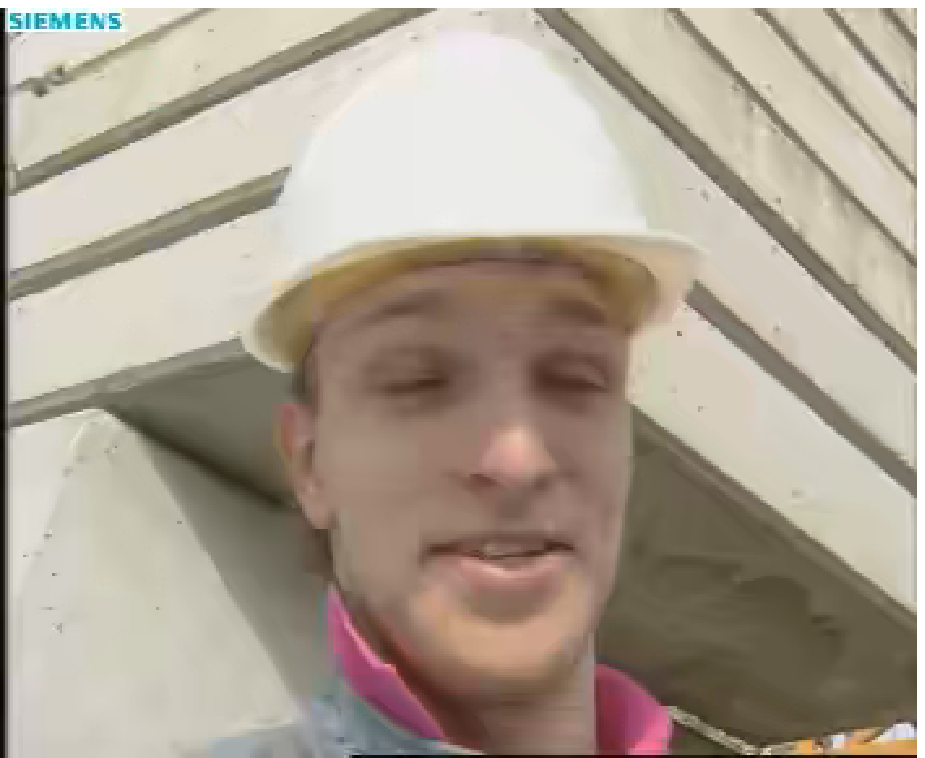}
\label{f:foreman44adds}}
\subfigure[Scenario (ii)]
{\includegraphics[width=0.25\textwidth]{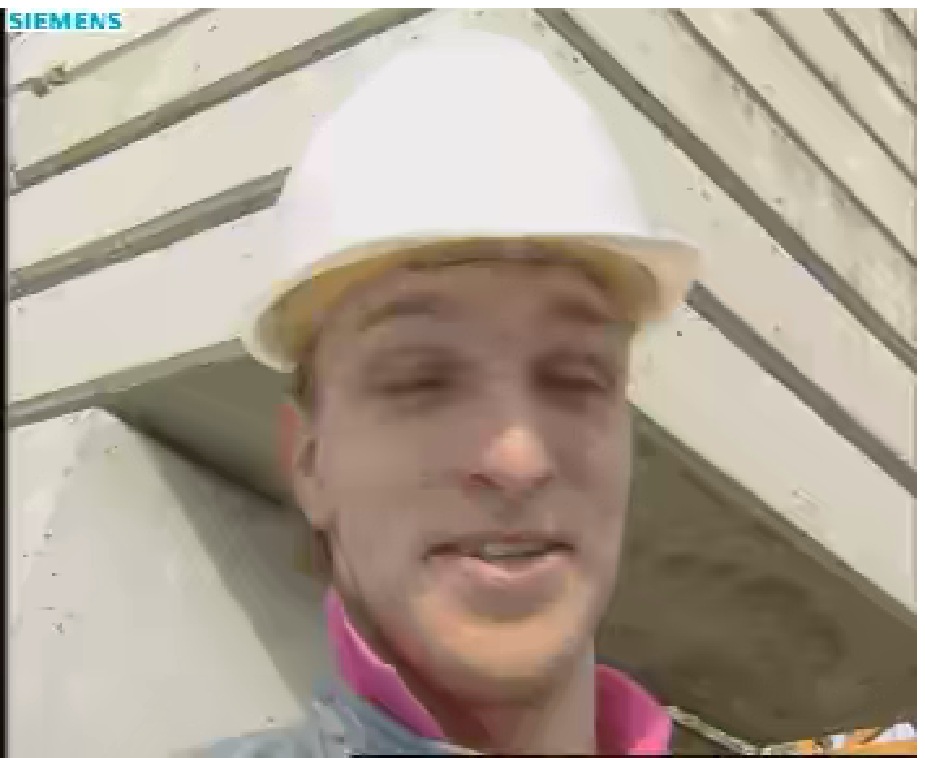}
\label{f:foreman14and44adds}}
\caption{First frame from `Foreman' video in the HEVC experiment with $\text{QP}=35$.
}
\label{fig:hevcframes}
\end{figure}

In both scenarios,
we have considered 11~CIF videos of 300~frames
obtained from a public video database~\cite{videos}.
The default HEVC coding
configuration for \texttt{Main} profile
was
adopted,
which includes both 8-bit depth intra and inter-frame coding modes.
We varied the quantization parameter (QP)
from 5 to 50 in steps of~5.
We adopted the PSNR as figure of merit,
because it is readily available in the reference software.
Measurements were taken for each color channel and frame.
The overall video PSNR value was computed according to~\cite{Ohm2012}.
Average PSNR measurements
are shown in Fig.~\ref{fig:graphicshevc}.
The proposed approximation
is multiplierless
and
effected
66\% and 53{.}12\% savings in the number of additions
considering
Scenarios (i) and (ii), respectively.
At the same time,
the resulting
image quality measures
showed
average errors
less than 0{.}28\% and 0{.}71\%,
for
Scenarios~(i) and~(ii), respectively.
Fig.~\ref{fig:hevcframes}
displays
the first frame of the Foreman encoded video
according to
the unmodified codec
and the modified codec in Scenarios~(i)~and~(ii).
The approximate transform
could effect
images that are essentially
identical to the ones produced
by the actual codec
at a much lower computational complexity.

\section{Hardware implementation}
\label{section-hardware}

In order to evaluate
the hardware resource consumption of
the proposed approximation,
it was modeled and tested
in Matlab Simulink and
then it was physically realized on
FPGA.
The employed FPGA was a Xilinx Virtex-6 XC6VLX240T
installed on a Xilinx ML605 prototyping board.
The FPGA realization was tested with 10,000 random
16-point input test vectors
using hardware co-simulation.
Test vectors were generated from within the Matlab environment
and routed to the physical FPGA device
using JTAG based hardware co-simulation.
Then the data measured from the FPGA was routed back to Matlab
memory space.

The
associated
FPGA implementation
was evaluated
for hardware complexity and real-time performance using metrics
such as
configurable logic blocks (CLB) and flip-flop (FF) count,
critical path delay ($T_\text{cpd}$) in~ns,
and maximum operating frequency ($F_\text{max}$) in~MHz.
Values were obtained from the Xilinx FPGA synthesis
and place-route tools by
accessing the \texttt{xflow.results} report file.
In addition,
the
dynamic power
($D_p$) in $\mathrm{mW}/\mathrm{GHz}$
and
static power consumption ($Q_p$) in $\mathrm{mW}$
were
estimated using the Xilinx XPower Analyzer.
Using the CLB count as a metric
to estimate the circuit area~($A$)
and
deriving time~($T$) from $T_\text{cpd}$,
we also report
area-time complexity~($AT$)
and
area-time-squared complexity~($AT^2$).

Because the transformation in~\cite{Jridi2015}
possesses a very low arithmetic complexity (cf.~Table~\ref{tab:complexity})
and presents good performance (cf.~Table~\ref{tab:performances}),
it was chosen for a direct comparison with the proposed approximation.
The obtained results are displayed in Table~\ref{FPGAresults}.
The proposed approximation presents
an improvement of 41.28\% and 43.26\%
in area-time and area-time-square measures,
respectively, when compared to~\cite{Jridi2015}.

\begin{table}
\centering
\caption{Hardware resource and power consumption using Xilinx Virtex-6 XC6VLX240T 1FFG1156 device}
\label{FPGAresults}
\begin{tabular}{lc@{\,\,}c@{\,\,}c@{\,\,}c@{\,\,}c@{\,\,}c@{\,\,}c@{\,\,}c}
\toprule
Method &
CLB &
FF &
$T_\text{cpd}$%
&
$F_{\text{max}}$%
&
$D_p$%
&
$Q_p$%
&
$AT$ &
$AT^2$\\
\midrule
Transform in~\cite{Jridi2015} & 499 & 1588 & 3.0 & 333.33 & 7.4 & 3.500 & 1497 & 4491\\
Proposed approx.              & 303 & 936  & 2.9 & 344.83 & 7.9 & 3.509 & 879  & 2548\\
\bottomrule
\end{tabular}
\end{table}

\section{Conclusion}\label{sec:conclusion}

This paper introduced
an orthogonal 16-point DCT approximation
which requires only 44 additions for its computation.
To the best of our knowledge,
the proposed transformation
has the \emph{lowest} computational cost
among the meaningful 16-point DCT approximations
archived in literature.
The introduced method
requires
from 26.67\% to 38.89\%
fewer arithmetic operations
than the best competitors.
In the context of image compression,
the proposed tool
attained
the best performance vs computational cost
ratio
for both PSNR and SSIM metrics.
When embedded into the H.265/HECV
standard,
resulting video frames
exhibited almost imperceptible degradation,
while demanding no multiplications
and 56~fewer additions
than the standard unmodified codec.
The hardware realization of the proposed transform presented
an improvement of more than 30\% in area-time and area-time-square measures
when compared to the lowest complexity competitor~\cite{Jridi2015}.
Potentially,
the present approach
can extended to derive 32- and 64-point approximations
by means of the scaled approach
introduced in~\cite{Jridi2015}.

\section*{Acknowledgments}

Authors acknowledge
CAPES, CNPq, FACEPE, and FAPERGS
for the partial support.

{\small
\bibliographystyle{IEEEtran}
\bibliography{bibcleaned}
}

\end{document}